\documentclass[conference]{IEEEtran}
\IEEEoverridecommandlockouts
\usepackage{cite}
\usepackage{amsmath,amssymb,amsfonts}
\usepackage{algorithmic}
\usepackage{graphicx}
\usepackage{textcomp}
\usepackage{xcolor}
\usepackage{graphics} 
\usepackage{epsfig} 
\usepackage{mathptmx} 
\usepackage{times} 
\usepackage{microtype}
\usepackage{algorithmic}
\usepackage{textcomp}
\usepackage{xcolor}
\usepackage{gensymb}
\usepackage{caption}
\usepackage{hyperref}
\usepackage{mwe}
\usepackage{verbatim}
\usepackage{amsbsy}
\usepackage{siunitx}
\usepackage{tabularx,booktabs}
\usepackage{dblfloatfix}
\usepackage{cite}
\usepackage{bm}
\usepackage{amsmath,amssymb,amsfonts}
\usepackage{subcaption}
\usepackage{float}

\usepackage{algorithm, algorithmic}
\usepackage[autolanguage]{numprint}

\usepackage{spreadtab}
\usepackage{multirow}
\usepackage{cite}
\usepackage{hyperref}
\usepackage{xcolor}
\usepackage{algorithmic}
\usepackage{textcomp}
\usepackage{xcolor}
\usepackage{gensymb}
\usepackage{caption}
\usepackage{mwe}
\usepackage{amsfonts}
\usepackage{amssymb}
\usepackage{verbatim}
\usepackage{amsbsy}
\usepackage{siunitx}
\usepackage{tabularx, booktabs}
\usepackage{soul}
\usepackage{tikz}
\usetikzlibrary{calc}
\def\BibTeX{{\rm B\kern-.05em{\sc i\kern-.025em b}\kern-.08em
    T\kern-.1667em\lower.7ex\hbox{E}\kern-.125emX}}

\makeatletter
\newif\if@anonymize

\@anonymizetrue    

\if@anonymize
  \newcommand{\highlight@DoHighlight}{
    \fill [outer sep = -15pt, inner sep = 0pt, color=black]
          ($(begin highlight)+(0,8pt)$) rectangle ($(end highlight)+(0,-3pt)$) ;
  }

  \newcommand{\highlight@BeginHighlight}{
    \coordinate (begin highlight) at (0,0) ;
  }

  \newcommand{\highlight@EndHighlight}{
    \coordinate (end highlight) at (0,0) ;
  }

  \newdimen\highlight@previous
  \newdimen\highlight@current
  \newlength{\item@width}

  \DeclareRobustCommand*\anonymize{%
    \SOUL@setup
    \def\SOUL@preamble{%
      \begin{tikzpicture}[overlay, remember picture]
        \highlight@BeginHighlight
        \highlight@EndHighlight
      \end{tikzpicture}%
    }%
    \def\SOUL@postamble{%
      \begin{tikzpicture}[overlay, remember picture]
        \highlight@EndHighlight
        \highlight@DoHighlight
      \end{tikzpicture}%
    }%
    \def\SOUL@everyhyphen{%
      \discretionary{%
        \SOUL@setkern\SOUL@hyphkern
        \SOUL@sethyphenchar
        \tikz[overlay, remember picture] \highlight@EndHighlight ;%
      }{%
      }{%
        \SOUL@setkern\SOUL@charkern
      }%
    }%
    \def\SOUL@everyexhyphen##1{%
      \SOUL@setkern\SOUL@hyphkern
      \settowidth{\item@width}{##1}%
      \makebox[\item@width]{}%
      \discretionary{%
        \tikz[overlay, remember picture] \highlight@EndHighlight ;%
      }{%
      }{%
        \SOUL@setkern\SOUL@charkern
      }%
    }%
    \def\SOUL@everysyllable{%
      \begin{tikzpicture}[overlay, remember picture]
        \path let \p0 = (begin highlight), \p1 = (0,0) in \pgfextra
          \global\highlight@previous=\y0
          \global\highlight@current =\y1
        \endpgfextra (0,0) ;
        \ifdim\highlight@current < \highlight@previous
          \highlight@DoHighlight
          \highlight@BeginHighlight
        \fi
      \end{tikzpicture}%
      \settowidth{\item@width}{\the\SOUL@syllable}%
      \makebox[\item@width]{}%
      \tikz[overlay, remember picture] \highlight@EndHighlight ;%
    }%
    \SOUL@
  }
\else
  \newcommand{\anonymize}[1]{#1}
\fi

\newcommand{\linebreakand}{%
  \end{@IEEEauthorhalign}
  \hfill\mbox{}\par
  \mbox{}\hfill\begin{@IEEEauthorhalign}
}

\makeatother    

\title{Behavior Cloning for Mini Autonomous Car Path Following}

\author{\IEEEauthorblockN{Pablo Moraes, Christopher Peters, Hiago Sodre, William Moraes,\\ Sebastian Barcelona, Juan Deniz, Victor Castelli, Bruna Guterres,\\ Ricardo Grando}
\IEEEauthorblockA{Technological University of Uruguay, UTEC, Uruguay}
\IEEEauthorblockA{Ostfalia University of Applied Sciences, Germany}
\IEEEauthorblockA{Corresponding Author: pablo.moraes@utec.edu.uy}
}

\begin{document}

\maketitle

\begin{abstract}
This article presents the implementation and evaluation of a behavior cloning approach for route following with autonomous cars. Behavior cloning is a machine-learning technique in which a neural network is trained to mimic the driving behavior of a human operator. Using camera data that captures the environment and the vehicle's movement, the neural network learns to predict the control actions necessary to follow a predetermined route. Mini-autonomous cars, which provide a good benchmark for use, are employed as a testing platform. This approach simplifies the control system by directly mapping the driver's movements to the control outputs, avoiding the need for complex algorithms. We performed an evaluation in a 13-meter sizer route, where our vehicle was evaluated. The results show that behavior cloning allows for a smooth and precise route, allowing it to be a full-sized vehicle and enabling an effective transition from small-scale experiments to real-world implementations.

\href{https://youtu.be/BBRmsOxTdRE}{Supplementary Video: https://youtu.be/BBRmsOxTdRE}.
\end{abstract}


\section{Introduction}

One of the main challenges with autonomous cars is achieving precise route following. Mini-autonomous cars are useful tools for testing and validating new algorithms because they are economical and easy to use. They feature sensors similar to those in larger autonomous vehicles, allowing for rapid and effective testing of various control systems \cite{kocic2018driver}.

Behavior cloning is a machine-learning technique used in autonomous driving \cite{le2022survey}. A neural network is trained to mimic the driving behavior of a human driver using camera data. The process involves a human driver operating the vehicle along several routes while the camera captures environmental information and the car's movement data. These data are used to train a neural network, which learns to predict control actions based on the vehicle and camera data collected \cite{le2022survey}.

Behavior cloning simplifies the control system by mapping the driver's movements to the control outputs, avoiding the need for complex algorithms. Behavior cloning can be scaled from mini-autonomous cars to full-sized vehicles, allowing for a seamless transition from small-scale experiments to real-world implementations \cite{satilmics2019cnn}. This article presents the implementation and evaluation of a behavior cloning approach for route following in mini-autonomous cars, demonstrating that this method can achieve smooth and precise route following.

In this work we propose the use of behavior cloning for mini autonomous car racing. Our methodology consists of using a convolutional neural network that learns based on a camera vision and  an user experience on a race track. We also perform an ablation study, providing insights on how the variation of the model impacts on the final vehicles' performance.

This paper is organized as follows: In the sequence, we present a Related Works Section (Sec. \ref{related_works}). Our methodology is presented in Sec. \ref{methodology}, followed by our evaluation Sec. \ref{results}. Our contributions and present future works are highlighted in Sec. \ref{conclusion}.

\section{Related Works}
\label{related_works}

Farag et al. \cite{farag2018behavior} proposed the use of a CNN for controlling autonomous cars. The model consisted of seventeen convolutional layers trained with a dataset of samples with frontal cameras and the steering commands generated by a driver on urban roads. The output neuron generated the steering command for the car.

Kocic et al. \cite{kocic2018driver} proposed a solution for driver behavioral cloning using deep learning to achieve autonomous driving in simulated conditions. The dataset was created in a simulator that had a camera in front of the vehicle, simulating the driver's vision. The behavioral cloning was based on the NVIDIA convolutional neural network model.

Le Mero et al. \cite{le2022survey} performed a review of behavior cloning for autonomous driving. It described behavior cloning and divided it into 3 paradigms: end-to-end control prediction, direct perception, and uncertainty quantification. The first one, end-to-end control prediction, was categorized as training a model to map from input data to signals of autonomous driving, like steering angle and throttle. This paradigm was used in the proposed methodology of this work.

Chu et al. \cite{chu2020sim} proposed a sim-to-real autonomous car system that relied on reinforcement learning instead of supervised learning. The model learned from the user experience, and with that, the authors concluded that a better performance could be achieved.

Overall, our methodology was inspired on the mentioned related works. We also perform behavior cloning using an convolutional neural network trained using supervisioned learning. We use two outputs that provides the cars steering and the throttle and our dataset was also created using a a real user experience driving the vehicle in a race track.

\section{Methodology}
\label{methodology}

\subsection{Vehicle Assembly}

\begin{figure}[ht]
    \centering
    \begin{minipage}[b]{0.2\textwidth} 
        \centering
        \begin{tabular}{|c|c|}
        \hline
        \textbf{Dimension} & \textbf{Size} \\
        \hline
        Length & 330 mm \\
        \hline
        Width & 190 mm \\
        \hline
        Height & 170 mm \\
        \hline
        Weight & 2.2 kg \\
        \hline
        \end{tabular}
        \caption{Specifications of the Autonomous Car}
        \label{tab:car_specifications}
    \end{minipage}%
    \hfill 
    \begin{minipage}[b]{0.2\textwidth} 
        \centering
        \includegraphics[width=0.9\linewidth]{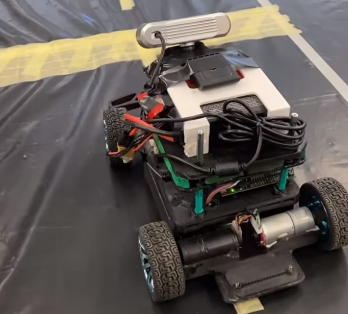}
        \caption{Autonomous car in evaluation environment}
        \label{fig:Autonomous}
    \end{minipage}
\end{figure}

To introduce the construction specifications of this vehicle, it is possible to view its actual physical dimensions Table \ref{tab:car_specifications}. The vehicle's hardware was fully developed according to the competition's needs, with fully defined dimensions. The description of its parts is included in the following subsection with the description of the components and their functions for operation. An initial image of the vehicle in its configuration phase for operation can be seen in Figure ~\ref{fig:Autonomous}.

Our autonomous vehicle is characterized by a design that combines PLA+ in its chassis to ensure both strength and lightness, and it also has a low-cost bias. Its computational core lies in a Jetson Nano 4GB, backed by a 3D-printed casing for protection. Equipped with a Logitech C920 HD Pro camera for real-time vision and powered by two 11.1V LiPo batteries, the vehicle ensures real-time performance.

The vehicle's locomotion is achieved through 6V DC motors and an SG5010 servo motor, controlled respectively by a PCA9685 and an L298N H-bridge. Additionally, a TP-Link TX20 USB WiFi adapter is incorporated for wireless connectivity. The vehicle uses 19mm bearings and soft iron shafts for the Ackerman system. An XL4016 step-down converter is also included to power the Jetson Nano. These components are interconnected following a specific scheme, ensuring coordinated and efficient operation throughout all autonomous vehicle operations.

\begin{table}[ht]
\centering
\caption{Components of our autonomous vehicle}
\label{tab:components}
\begin{tabular}{|c|c|c|}
\hline
  \includegraphics[width=0.08\textwidth]{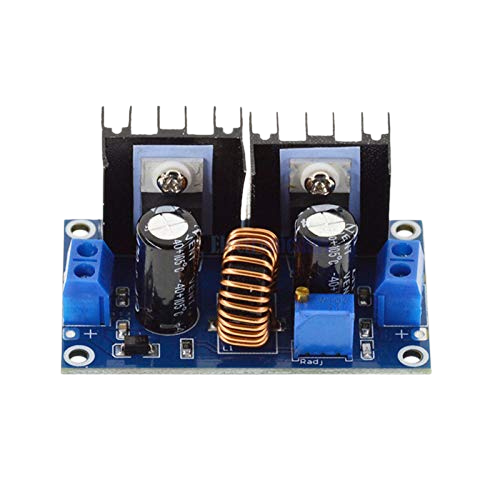} & 
  \includegraphics[width=0.08\textwidth]{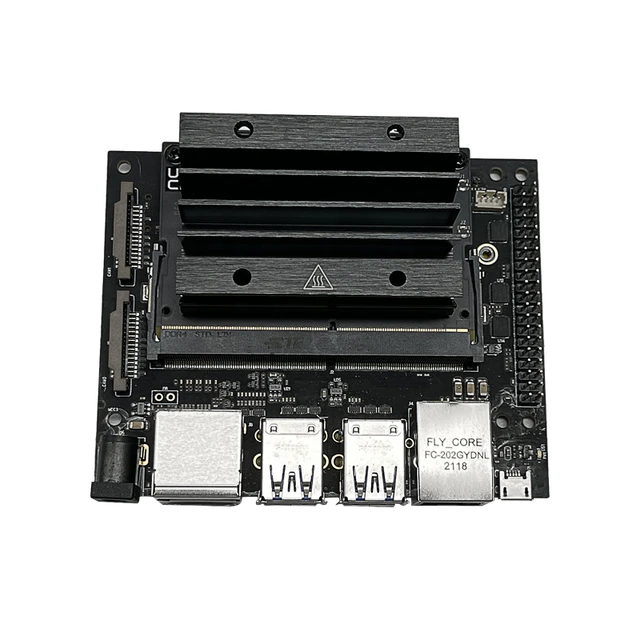} &
  \includegraphics[width=0.08\textwidth]{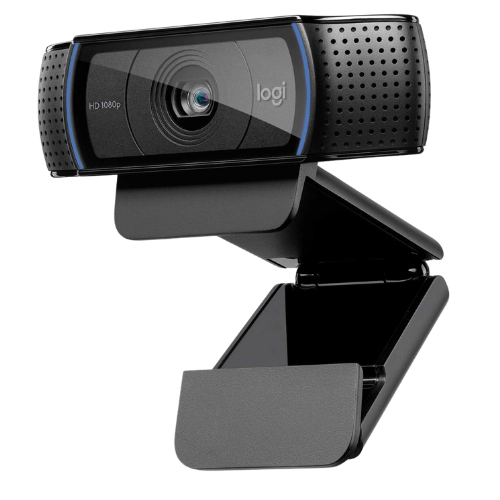} \\
  (a) & (b) & (c) \\
\hline
  \includegraphics[width=0.08\textwidth]{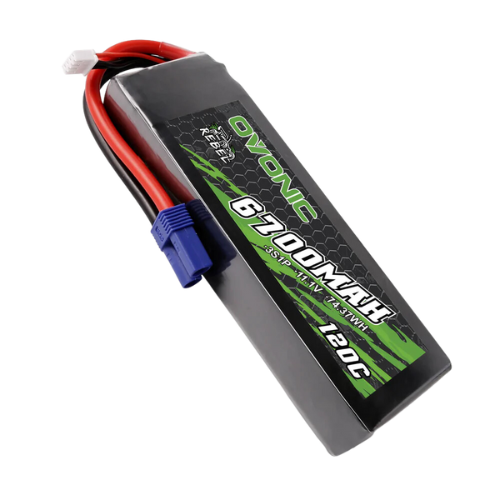} & 
  \includegraphics[width=0.08\textwidth]{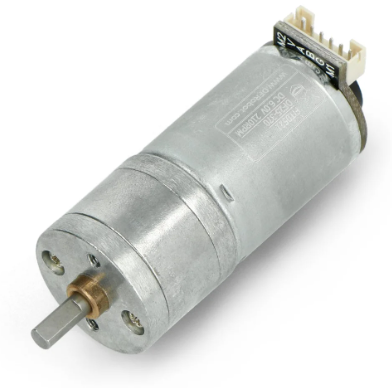} &
  \includegraphics[width=0.08\textwidth]{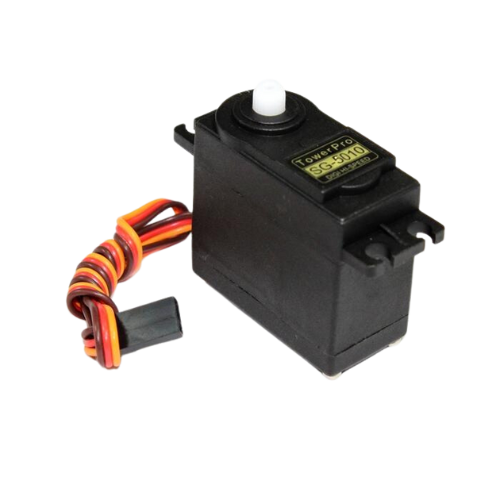} \\
  (d) & (e) & (f) \\
\hline
  \includegraphics[width=0.08\textwidth]{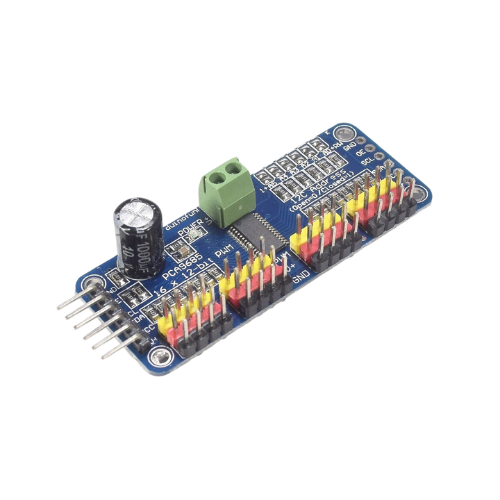} & 
  \includegraphics[width=0.08\textwidth]{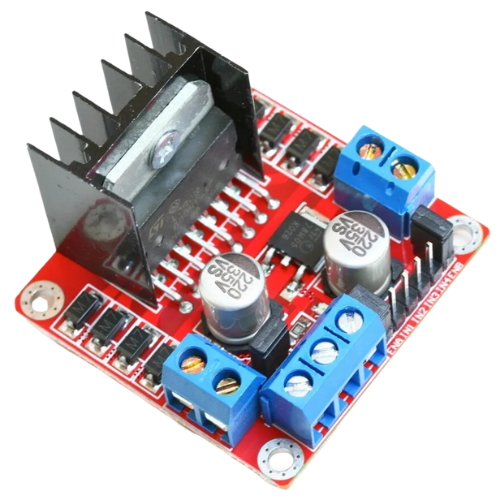} &
  \includegraphics[width=0.08\textwidth]{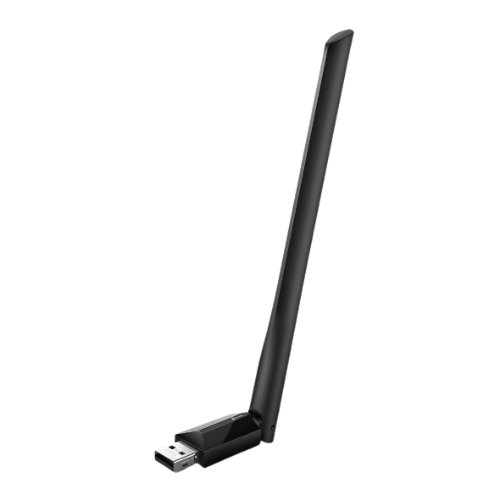} \\
  (g) & (h) & (i) \\
\hline

\end{tabular}
\end{table}

\begin{table}[ht]
    \centering
    \caption{Presented below is a tabulated enumeration of the components delineated within.}
    \label{tab:component_details}
    \renewcommand{\arraystretch}{1.3} 
    \resizebox{\columnwidth}{!}{ 
    \begin{tabular}{|p{3cm}|p{5cm}|} 
    \hline 
    \textbf{Component} & \textbf{Details} \\ 
    \hline 
    Jetson Nano 4GB & A compact, powerful single-board computer by NVIDIA with a quad-core ARM CPU, 128-core GPU, and 4GB RAM (see Figure b).\\ 
    \hline 
    LiPo Battery 11.1V 3S & These are 11.1V 3-cell lithium polymer batteries, commonly used in RC vehicles, drones, and other high-power, lightweight electronic devices (see Figure d).\\ 
    \hline 
    Logitech C920 & The Logitech C920 is a high-definition webcam known for its sharp video and audio quality (see Figure c).\\ 
    \hline 
    DC Motors 6V & These 6-volt DC motors are versatile and are used in robotics applications for their simplicity and torque-boosting system (see Figure e).\\ 
    \hline 
    PCA 9685 & The PCA9685 is a 16-channel, 12-bit PWM controller used for precise control of multiple servo motors and DC motors, connected to our H-bridge (see Figure g).\\ 
    \hline 
    L298N Bridge & An H-Bridge is a circuit that allows control of the direction and speed of DC motors (see Figure h).\\ 
    \hline 
    XL4016 Buck Converter & The XL4016 is a DC-DC converter that reduces a high input voltage to a lower one. It is used to stably power our Jetson Nano (see Figure a).\\ 
    \hline 
    SG5010 Servomotor & The SG5010 is a servo motor known for being affordable and useful in robotics. It has very precise positioning control (see Figure f).\\ 
    \hline 
    TP-LINK TX20u USB Adapter & The TP-LINK TX20u USB Adapter enables our Jetson Nano, which lacks built-in Wi-Fi, to connect to wireless networks (see Figure i).\\ 
    \hline
    \end{tabular}
    }
\end{table}

\subsection{Convolutional Neural Network}

A Convolutional Neural Network (CNN) designed to work on devices such as the Jetson Nano was used. The network has five convolutional layers that help identify and understand the images captured by the vehicle camera. We used 3x3 filters on these layers and enabled each filter with a function called ReLU so that the image processing retains important variations. We adjusted the model by starting with 24 filters in the first layer and increasing the number to 64 filters in the last layer. This helps capture more detail as we move through the layers.

After the convolutional layers, we add dense layers that take all the information from the images and decide how the vehicle should move, both in speed and direction. To prevent the model from fitting too closely to our dataset and not performing well in real situations, we incorporate dropout layers. These layers basically 'turn off' some random neurons during training.

The model is trained with a set of images that fits a standard size of 120x160 pixels. This set of images, varies depending on the dataset created, the images come labeled with the correct steering and acceleration actions, allowing to teach the model how to respond to what it sees on the track. This method allows our vehicle to be handled well on the track, adapting to different situations and reacting to what the camera sees in front of it.

\begin{figure}[ht]
    \centering
    \includegraphics[width=0.65\linewidth]{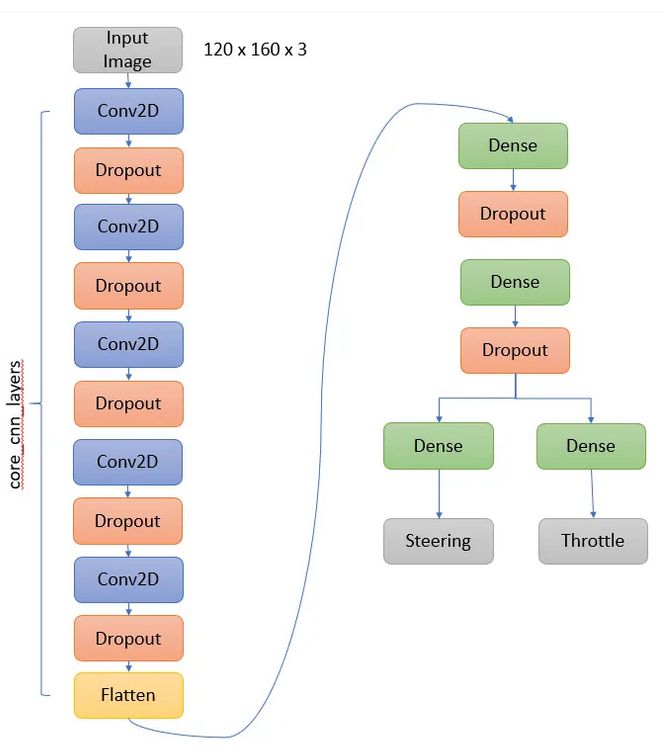}
    \captionsetup{justification=centering}
    \caption{Network Structure}
    \label{fig:CNN}
\end{figure}

Here is the schematic of the CNN architecture used in our autonomous vehicle Fig.~\ref{fig:CNN}, showing the sequence of convolutional layers, dropout layers and dense layers culminating in the vehicle's steering and acceleration controls.

\subsection{Evaluation Environment}

To verify the effectiveness of our convolutional neural network (CNN) in behavior cloning for an autonomous car, we created an environment that simulated real track conditions. The designed circuit included some curves to test the vehicle in various driving situations. The circuit measures 3x3 meters, ideal for assessing how the vehicle follows a predetermined path and how it handles changes in direction.

During the tests, we collected some data on the vehicle's performance, such as speed and direction, using cameras installed around our circuit. This data was important for analyzing the autonomous car's ability to respond to different CNN configurations. We observed how the vehicle executed each lap, paying attention to how it handled the curves and maintained the desired trajectory. This monitoring allowed us to verify the accuracy of the route tracking and also observe its response to imperfections in the circuit.

The results were documented and comparison tables and performance metrics were generated, which will be discussed in the following sections. These metrics include lap times and deviations from the ideal trajectory, as well as other observations that provide information on the vehicle's behavior using different neural network configurations.

\section{Results}
\label{results}

In this section, we present the results obtained from three tests conducted on the autonomous vehicle with three different configurations of our neural network. Each model was evaluated under the same conditions on a circuit simulating a real environment.

\begin{figure*}[!ht]
    \centering
    \subfloat[Original Model.]{\includegraphics[width=0.3\linewidth]{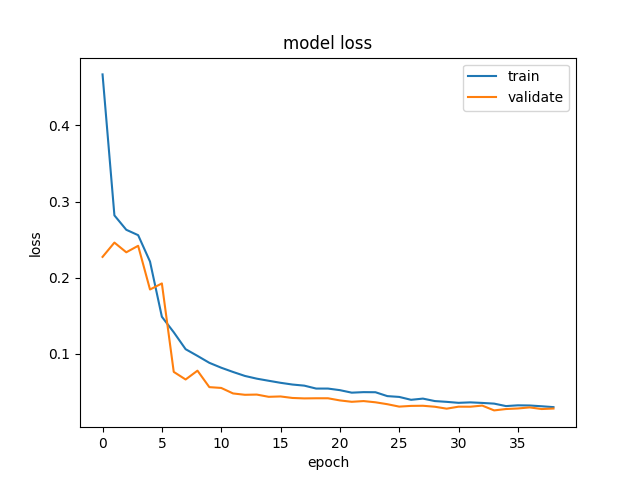}\label{fig:modelo_original}}
    \subfloat[Model with One Less Layer.]{\includegraphics[width=0.3\linewidth]{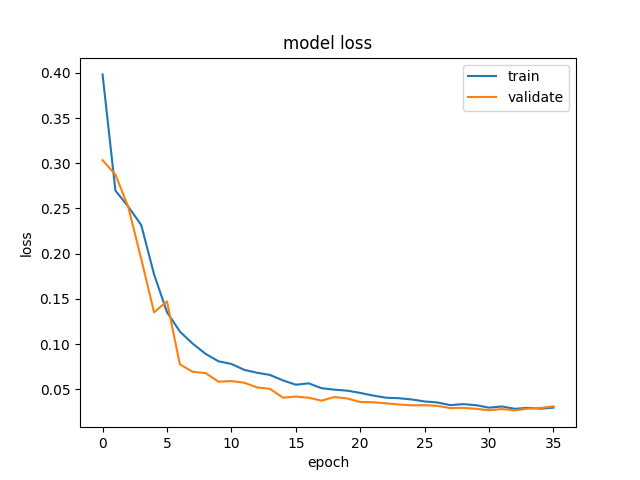}\label{fig:modelo_capa_menos}}
    \subfloat[Model with an Additional Layer.]{\includegraphics[width=0.3\linewidth]{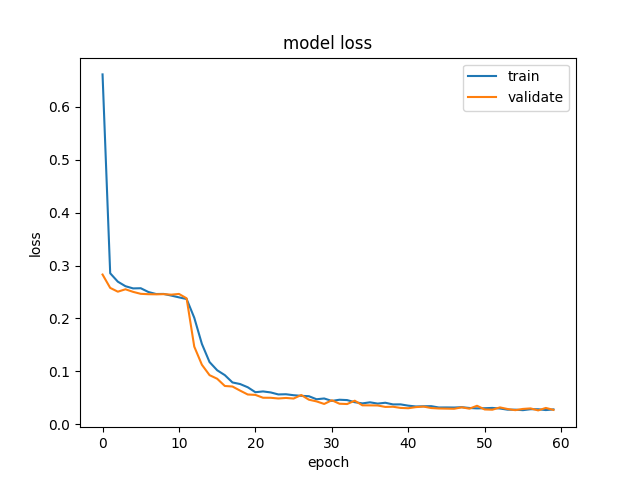}\label{fig:modelo_capa_mas}}    
    \caption{Loss of the CNN models used for evaluation.}
    \label{fig:environment1}
    \vspace{-5mm}
\end{figure*}

We used a dataset with \textbf{3631} images. The training of the first model ended with \textbf{817028} parameters and took \textbf{43979} microseconds, as shown in Fig. \ref{fig:modelo_original}. \\
The second model had \textbf{1074500} parameters and \textbf{43208} microseconds, as shown in Fig. \ref{fig:modelo_capa_menos}. \\
The third model had \textbf{610756} parameters and \textbf{49874} microseconds, as shown in Fig. \ref{fig:modelo_capa_mas}.

\subsection{Test Results}

\textbf{Original Model:}
\begin{table}[htbp]
\centering
\caption{Lap Results for the Original Model}
\label{tab:model_original}
\begin{tabular}{|c|c|c|}
\hline
\textbf{Lap} & \textbf{Time (seconds)} & \textbf{Off Track} \\ \hline
1 & 22.63 & No \\ \hline
2 & 18.65 & Almost Went Off  \\ \hline
3 & 20.51 & No \\ \hline
4 & 20.34 & No \\ \hline
5 & 20.43 & No \\ \hline
6 & 20.56 & Almost Went Off  \\ \hline
7 & 20.45 & No \\ \hline
8 & 20.70 & Goes Off Track \\ \hline
9 & 18.17 & Almost Went Off \\ \hline
10 & 20.70 & Sí \\ \hline
\end{tabular}
\end{table}

\textbf{Model with One Less Layer:}
\begin{table}[htbp]
\centering
\caption{Lap Results for the Model with One Less Layer}
\label{tab:model_less_layer}
\begin{tabular}{|c|c|c|}
\hline
\textbf{Lap} & \textbf{Time(seconds)} & \textbf{Off Track} \\ \hline
1 & 21.70 & Sí \\ \hline
2 & 20.51 & No \\ \hline
3 & 20.46 & No \\ \hline
4 & 20.46 & No \\ \hline
5 & 18.90 & Goes Off Track\\ \hline
6 & 21.37 & No \\ \hline
7 & 20.28 & No \\ \hline
8 & 20.13 & Almost Went Off \\ \hline
9 & 20.27 & No \\ \hline
10 & 19.97 & No \\ \hline
\end{tabular}
\end{table}

\textbf{Model with an Additional Layer:}
\begin{table}[htbp]
\centering
\caption{Lap Results for the Model with an Additional Layer}
\label{tab:model_more_layer}
\begin{tabular}{|c|c|c|}
\hline
\textbf{Lap} & \textbf{Time (seconds)} & \textbf{Off Track} \\ \hline
1 & 22.24 & Casi sale \\ \hline
2 & 20.60 & No \\ \hline
3 & 21.51 & Sale y vuelve \\ \hline
4 & 21.71 & No \\ \hline
5 & 20.88 & No \\ \hline
6 & 21.05 & No \\ \hline
7 & 20.78 & Almost Went Off  \\ \hline
8 & 21.37 & Goes Off and Returns\\ \hline
9 & 21.36 & Goes Off and Returns\\ \hline
10 & 21.33 & No \\ \hline
\end{tabular}
\end{table}

\subsection{Model Comparison}

\begin{table}[htbp]
\centering
\caption{Comparison of Lap Results}
\label{tab:model_comparison}
\begin{tabular}{|c|c|c|c|}
\hline
\textbf{Model} & \textbf{Average Time(s)} & \textbf{Deviations} & \textbf{Off Track Incidents} \\ \hline
Original & 20.35 & Low & 4 times \\ \hline
Less Layer & 20.28 & Low & 2 times \\ \hline
More Layer & 21.03 & Medium & 5 times \\ \hline
\end{tabular}
\end{table} 

We compared how each model handled the circuit. We evaluated the training time required for each neural network configuration, which is important to understand the efficiency of each model in terms of learning and optimization.

\subsection{Observations}

The original model was the most precise in following the central line of the track, although it had four off-track incidents. The model with one less layer, despite having good overall performance and only two off-track incidents, experienced significant variations, switching between sides of the track, which made it more unstable.

The model with an additional layer was slower and had a higher number of deviations and off-track incidents. This model showed greater sensitivity to imperfections in the track, making maneuvering decisions in response to small details and obstacles detected along the way, which could explain its frequent course corrections.

These varied results suggest that while the original model offers the best accuracy on the track, modifications in the network architecture (both reduction and addition of layers) can significantly influence the vehicle's behavior, affecting the car's ability to maintain a stable and predictable trajectory, which can be utilized for other applications.

\begin{figure}[ht]
\centering
\includegraphics[scale=0.4]{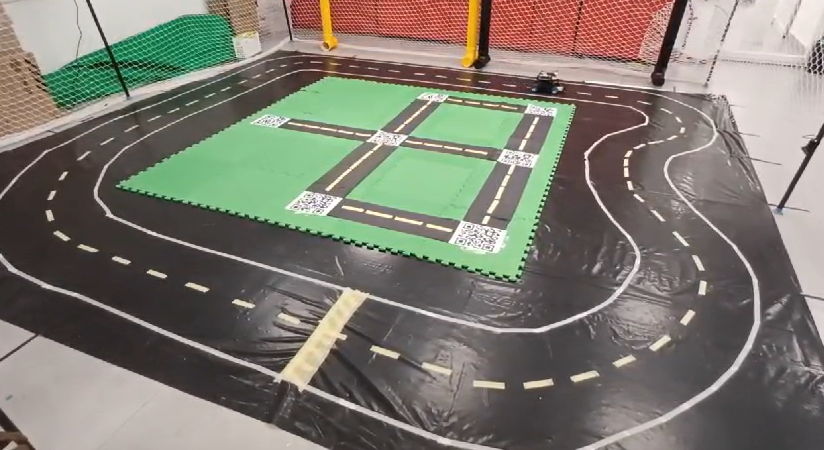}
\caption{Track Scenario Used to Validate Our Vehicle.}
\label{fig:scenario}
\end{figure}

\section{Conclusion}
\label{conclusion}

In this study, the effects of various convolutional neural network configurations on behavior cloning in mini autonomous cars were explored. The results demonstrated how variations in the network architecture impacted the vehicle's behavior. It was shown that a balance in the model and accuracy in cloning the environment is crucial.

For future work, we will investigate hybrid models and the application of reinforcement learning techniques to adjust the network parameters based on the conditions in which the vehicle operates. This adaptability could optimize the vehicle's efficiency in varied environments. Additionally, extending this research to larger vehicles would provide new insights into the scalability and applicability of behavior cloning techniques under more realistic conditions.

\section*{Acknowledgment}


The authors would like to thank the Technological University of Uruguay team and the UruBots robotics competitions team in Rivera.

\bibliographystyle{IEEEtran}
\bibliography{bibliography/IEEEexample}

\end{document}